\documentclass[10pt,twocolumn,letterpaper]{article}

\usepackage{cvpr}
\usepackage{times}
\usepackage{epsfig}
\usepackage{graphicx}
\usepackage{amsmath}
\usepackage{amssymb}
\usepackage{colortbl}
\usepackage{arydshln}
\usepackage{tabularx}
\usepackage{comment}
\definecolor{Gray}{gray}{0.9}
  \pdfoutput=1

\usepackage{xspace}
\makeatletter
\DeclareRobustCommand\onedot{\futurelet\@let@token\@onedot}
\def\@onedot{\ifx\@let@token.\else.\null\fi\xspace}

\def\eg{\emph{e.g}\onedot} 
\def\ie{\emph{i.e}\onedot}

\makeatother

\usepackage[pagebackref=true,breaklinks=true,colorlinks,bookmarks=false]{hyperref}

 \cvprfinalcopy 

\def\cvprPaperID{4} 

\ifcvprfinal\fi

\begin{document}

\title{Textual Visual Semantic Dataset for Text Spotting}




\author{Ahmed Sabir$^{1}$ \hspace{0.7cm}
Francesc  Moreno-Noguer$^{2}$ \hspace{0.7cm}
 Llu\'{\i}s Padr\'o$^{1}$ \\
$^{1}$ Universitat Polit\`ecnica de Catalunya, TALP Research Center, Barcelona, Spain  \\ 
$^{2}$ Institut de Rob\`otica i Inform\`atica Industrial, CSIC-UPC, Barcelona, Spain\\
}

\maketitle

\begin{abstract}
Text Spotting in the wild consists of detecting and recognizing text appearing in  images (e.g.  signboards, traffic signals or brands in clothing or objects). This is a challenging problem due to the complexity of the context where texts appear (uneven backgrounds, shading, occlusions, perspective distortions, etc.). Only a few  approaches try to exploit the relation between text and its surrounding environment to better recognize text in the scene. In this paper, we propose a visual context dataset\footnote{Our dataset is publicly available at: \href{https://git.io/JeZTb}{https://git.io/JeZTb}} for Text Spotting in the wild, where the  publicly available dataset COCO-text \cite{veit2016coco} has been extended with information about the scene (such as objects and places appearing in the image) to enable researchers to include semantic relations between texts and scene in their Text Spotting systems, and to offer a common framework for such approaches. For each text in an image, we extract three kinds of context information: objects in the scene, image location label and a textual image description (caption). We use state-of-the-art out-of-the-box available tools to extract this additional information. Since this information has textual form, it can be used to leverage text similarity or semantic relation methods into Text Spotting systems, either as a post-processing or in an end-to-end training strategy.
\end{abstract}

\section{Introduction}

Recognition of scene text in  images in the wild  is still an open problem in computer vision. There exist a number of difficulties in recognizing texts in images due to the many possible lighting conditions, variations in textures, complex backgrounds, textual font types and perspective distortions. The ability to automatically detect and recognize  text in natural images, a.k.a. \textit{text spotting or OCR in the wild} is an important challenge for many applications such as visually-impaired assistants \cite{liambas2016autonomous} or autonomous vehicles \cite{priambada2017levensthein}. In recent years, the interest of Computer Vision community in Text Spotting has significantly increased \cite{bissacco2013photoocr,shi2016end,jaderberg2016reading,ghosh2017visual,fang2018attention,Ramisa_pami2017}. However, state-of-the-art scene text recognition methods do not leverage object and scene recognition. Therefore, in this work, we introduce a visual semantic context textual dataset (\eg object, scene information) for text spotting tasks. Our goal is to fill this gap, bringing closer vision and language, by understanding the scene text and its relationship with the  environmental visual context. 

 \begin{figure}[t!]
 \centering 
\includegraphics[width=3.4in]{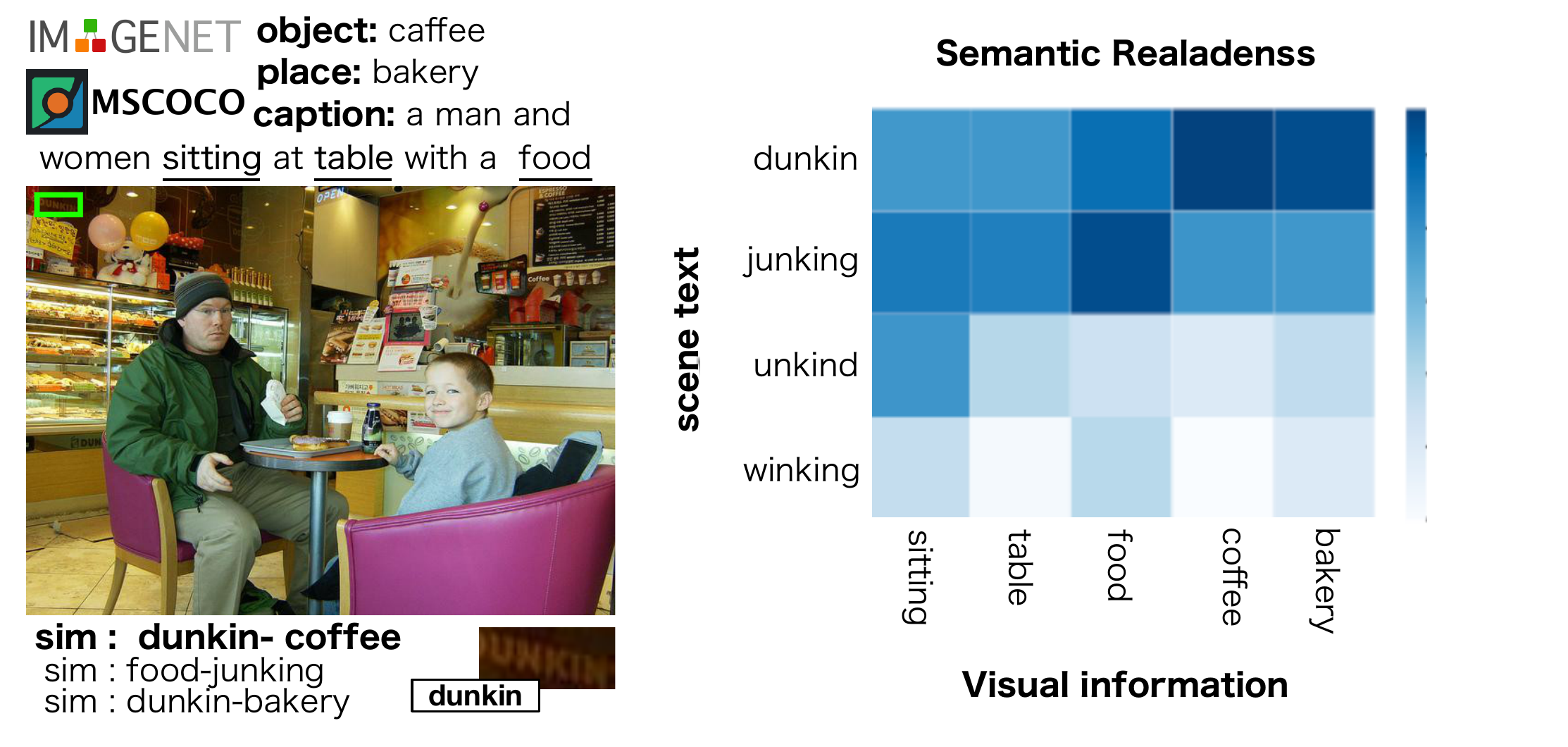}  
 \caption{Overview of the natural language understanding visual context information. The word ``dunkin" has a strong semantic relation with ``bakery", ``food", and ``coffee", thus it will be more likely to appear in this image than other similar words such as ``junking" or ``unkind". Note that this relation is computed by pre-traind word2vec \cite{mikolov2013distributed} cosine similarity. } 
 \label{fig:dount} 
 \end{figure}

The relation between text and its surrounding environment is very important to understand text in the scene. While there are some publicly available datasets for text spotting, none includes information about visual context in the image. Therefore, we propose a visual context semantic knowledge dataset for the text spotting pipeline, as our aim is to combine natural language processing and computer vision.  In particular, we exploit the semantic relatedness between the spotted text and its image context. For example, as shown in Figure \ref{fig:dount} the word ``dunkin" has a stronger semantic relation with ``coffee", thus it will be more likely to appear in a visual context than other possible candidates such as ``junking" or ``unkind".

 \begin{table*}
\caption{Several word recognition datasets. The images are cropped word images useful only for the recognition task only.}
\centering
\begin{tabular}{|l|l|c|c|c|}

\hline  
\multicolumn{5}{|c|}{ Text Recognition Dataset }  \\
\hline \hline  \rowcolor{Gray}
Label     &  Description  & Dictionary & \# bbox &  \# text \small $\dagger$ \\

\hline    \hline 
    IC17-T2 &  \  from COCO-text dataset ICDAR 17 Task-2 \cite{gomez2017icdar2017}              &  \  -  & \  46K & -\\

\hline 
 Synth90K   &  \ Synthetic dataset with 90K dict \cite{jaderberg2014synthetic}   & \  90K & \  9M   & -\\
\hline 
SVT        &  \  Street View Text \cite{wang2010word}  & \ - & \ 647 & - \\ 
\hline 
IC13       & \  ICDAR 2013 \cite{karatzas2013icdar}        &     -   &   \  1K & -  \\ 
\hline
\hline 
IC17-V &  \  Image+Textual dataset from IC17 Task-3 (ours)          &  \  -  & \  10K & 25K\\ 
\hline 
COCO-Text-V &  \   Image+Textual dataset from COCO-text       (ours)              &  \  -  & \  16K & 60K \\ 
\hline 
COCO-Pairs &  \   Only Textual dataset from COCO-text     (ours)              &  \  -  & \  - & 158K \\ 

\hline 

\end{tabular}
\vskip 0.01in
\small $\dagger$ Each sample have bounding box and its full image and the textual visual context (object, scene and caption). \\
\label{tb:tex-recogition}
\end{table*}

Departing from \cite{sabir2018visual,sabir2019semantic}, in this paper we describe in depth the construction of the visual context dataset. This dataset is based on the COCO-text \cite{veit2016coco}, which uses Microsoft COCO dataset \cite{lin2014microsoft} and annotates texts appearing in the images. We further extend the dataset using out-of-the-box tools to extract visual context or additional information from images.  Our main contribution is this combined visual context dataset, that provides the unrestricted-OCR research community the chance to use semantic relatedness between text and image to improve the results. The computer vision community tackles this problem by dividing the task into two sub-models one for text, and other for object \cite{zhu2016could,kang2017detection,Prasad_2018_ECCV}. 
Our approach uses existing state-of-the-art visual context generation approaches and thus, it can be used as a visual context with text spotting  as post-processing (OCR correction) or end-to-end training.

\section{Related Work}
While there are some publicly available datasets for text spotting, none of them includes visual
context information such as  objects in the scene, location id or  textual image descriptions.  In this section we describe several publicly available text spotting datasets. 

\subsection{Synthetic Dataset}
Table \ref{tb:tex-recogition} summarizes the number of examples in different datasets. Sizes of real image datasets with annotated texts are in the order of thousands, and have a very limited vocabulary, which makes them insufficient for deep learning methods. Therefore, \cite{jaderberg2014synthetic} introduced a synthetic data generator without any human label cost. Words are sampled from a 90K-words dictionary, and are rendered synthetically to generate images with complex background, fonts, distortions, etc. It contains 9 million cropped word text images. 
All current state-of-the-art text spotting algorithms~\cite{shi2016end,jaderberg2016reading,ghosh2017visual,fang2018attention} are trained on this dataset.

\subsection{ICDAR Dataset}

Text Spotting shared tasks carried out at ICDAR conferences released several relevant datasets:
\vspace{1mm}

\noindent{\bf ICDAR 2013 (IC13) \cite{karatzas2013icdar}.}
The ICDAR 2013 dataset consists of two sections for different text spotting subtasks: (1) text localization and (2) text segmentation. Text localization consists of 328 training images and 233 test images. Given its reduced size, ICDAR 2013 dataset is typically used for evaluation of scene text understanding tasks: localization, segmentation, and recognition.


\vspace{1mm}
\noindent{\bf ICDAR 2017 (IC17) \cite{gomez2017icdar2017}.}
ICDAR 2017 is based on COCO-text \cite{veit2016coco} and aims for end-to-end text spotting (\ie detection and recognition). The dataset consists of 43,686 full images with 145,859 text instances for training, and 10,000 images and 27,550 instances for validation.
\vspace{1mm}

\vspace{1mm}
\noindent{\bf Street View Text (SVT) \cite{wang2010word}.}
This  dataset consists of 349 images downloaded from Google Street View. For each image, only one word-level bounding box is provided. This is the first dataset that deals with text image in real scenarios, such as shop signs in a wide range of fonts styles. 

\vspace{1mm}
\noindent{\bf COCO-Text \cite{veit2016coco}.}
\label{sec:coco-text-explain}
This dataset is based on Microsoft COCO \cite{lin2014microsoft} (Common Objects in Context) and consists of 63,686 images, 173,589 text instance (annotations of the images). The COCO-text dataset differs from the other datasets in three aspects. First, the dataset was not collected with text recognition in mind. Thus, the annotated text instances lie in their natural context. Second, it contains a wide variety of text instances, such as machine-printed and handwritten text. Finally, the COCO-text has a much larger scale than other datasets for text detection and recognition. 

\section{Source Data}
We use state-of-the-art tools to extract textual information for each image. In particular, for each image we extract: 1) spotted text candidates (text hypotheses), and 2) surrounding visual context information.

\subsection{Text Hypotheses Extraction}
To extract the text associated with each image or bounding box we employ several off the self pre-trained Text Spotting baselines to generate $k$ text hypotheses. All the pre-trained models are trained on a synthetic dataset \cite{jaderberg2014synthetic}. 
We build out text hypotheses dataset for each image as the union of the predictions of all baselines. We next describe these models.


\vspace{1mm}
\noindent{\bf Convolutional Neural Network-90K-Dictionary  \cite{jaderberg2016reading}.}
The first baseline is a CNN  with fixed lexicon based recognition, able to recognize words in a predefined 90K-word dictionary $W$. Each word $w$ is corresponds to a word (class) in the 90K dictionary $W$ (multi-class classification). The dictionary is composed of various forms of English words (\eg nouns, verbs, adjectives, adverbs, etc) In short, the model classifies each input word-image into a pre-defined word in the 90K fixed lexicon. Each word $w\in W$ in the dictionary corresponds to one output neuron. The final output $word$ for a given image $x$ is written as:  
 \begin{equation}
word =\arg \max _{w \in \mathcal{W}} P(w |x, lexicon)
\end{equation}

\begin{table*}[t!]
\centering
\caption{Sample from the dataset. The text hypothesis comes from  existing Text Spotting baselines and the visual context information comes from out-of-the-box computer vision classifiers. The \textbf{bold} font is the ground-truth.} 
\begin{tabular}{|c|c|c|l|}

\hline 
  Text hypothesis  & Object & Scene & Caption  \\
  \hline 
  \hline 
 \textbf{11}, il, j, m, ...   & railroad & train & a train is on a train track with a train on it \\ 
 
lossing, docile, dow, \textbf{dell}, ...  & bookshop  & bookstore & a woman sitting at a table with a laptop \\ 

29th, 2th, \textbf{2011}, zit, ...& parking & shopping & a man is holding a cell phone while standing \\ 
\textbf{happy},	hooping, happily, nappy, ... & childs &  bib & a cake with a bunch of different types of scissors \\ 

\textbf{coke}, gulp, slurp, fluky,...    &  plate & pizzeria & a table with a pizza and a fork on it  \\ 
will, \textbf{wii}, xviii, wit,....     & remote  &  room & a close up of a remote control on a table  \\ 
 \hline 
\end{tabular}
\label{se:data}
\end{table*}

\vspace{1mm}
\noindent{\bf Convolutional Recurrent Neural Network (CRNN) \cite{shi2016end}.} The second baseline is a CRNN that learns the words directly from sequence labels, without relying on character annotations. The encoder uses a CNN to extract a set of features from the image. The CNN has no fully connected layers and extracts sequential feature representations of the input image, which are fed into a bidirectional RNN. A Connectionist Temporal Classification \cite{graves2006connectionist} based method is used to convert the per-frame predictions made by the RNN into a label sequence as following: 
\begin{equation}
\mathbf{I}^{*} \approx \mathcal{B}\left(\arg \max _{\pi} p(\boldsymbol{\pi} | \mathbf{y})\right)
\end{equation}
where $\mathcal{B}$ is sequence-to-sequence mapping function, $\boldsymbol{\pi}$ sequence label and $\mathbf{I}^{*}$ the sequence.

\vspace{1mm}
\noindent{\bf LSTM-Visual Attention (LSTM-V) \cite{ghosh2017visual}.}
The third baseline also generates output words as probable character sequences, without relying on lexicon. The network is based on encoder-decoder architecture with visual attention mechanism. In particular, they use the CNN pre-trained model \cite{jaderberg2016reading} mentioned above as encoder, but without the final layer, to extract the most important feature vectors from each text image. That feature vector is used to reduce model complexity through the soft attention model \cite{xu2015show}, which focus on the most relevant parts of the image at every step. The LSTM \cite{hochreiter1997long} decoder computes the output character probability $y$ for each time step and the visual attention $\alpha$, where $\mathrm{L}$ are the deep model output parameters of each layer:
\begin{equation}
P\left(y_{t} | \alpha , y_{t-1}\right) \sim \exp \left(\mathrm{L}_{0}\left(E y_{t-1}+\mathrm{L}_{\mathrm{h}} h_{t}+\mathrm{L}_{\mathrm{z}} \hat{z_{t}}\right)\right)
\end{equation}

\noindent{\bf CNN-Attention \cite{fang2018attention}.}
Finally, we employ one of  the most recent state-of-the art systems, which also produces the final output words as probable character sequences, without any fixed lexicon. The model is based on a CNN encoder-decoder  with attention and a CNN character language model. The final character prediction is a element-wise addition of the attention and language vector as:
\begin{equation}
p\left(y_{k} | y_{k-1}, \ldots, y_{1}\right)=p_{a}+p_{l} \\ 
\end{equation}
where $p_{a}$ and $p_{l}$ are softmax functions that convert the attention and language vectors to predicted characters separately.

\subsection{Visual Context Information}
To extract the \textit{visual context} from each image, we use out-of-the-box state-of-the-art classifiers. We obtain three kinds of contextual information: objects in the image, location/scenario labels, and a textual description or caption.

\subsubsection{Object information}
The output of the following classifiers is a $1000$-dimensional vector with the probabilities of $1000$ object types. We retain the top-5 most likely objects.

\vspace{1mm}
\noindent{\bf GoogLeNet \cite{szegedy2015going}. }
The design of this network  is based on an inception module, which uses 1-D convolutions to reduce the number of parameters. Also, a fully connected layer is replaced with a global average pooling at the end of the network. The network consists of 22 layer Deep CNN with reduced parameters. It has a top-5 error rate of 6.67\%.

\vspace{1mm}
\noindent{\bf Inception-ResNet \cite{szegedy2017inception}. }
Inspired by the  breakthrough ResNet performance, a hybrid-inception module was proposed. Inception-ResNet combines the two architectures (Inception modules and residual connection) to boost performance even further. We use Inception-ResNet-v2, with a top-5 error rate  3.1\%. 

The object hypotheses are obtained by extracting top-5 error class labels from each classifier and  re-ranking them based on the cosine distance.

\subsubsection{Scene information}
To extract scene information, we considered just one scene classifier \cite{zhou2017places}. This is a pre-trained scene classifier able to recognize $365$ different scenario classes. The original model is based on Places365-Standard as deep convolutions network that trained on 1.8 million images from 365 scene categories. The same work proposed a better model, which we use, consisting of a fine-tuned model {\textit{Places365-ResNet}}\footnote{\href{http://places2.csail.mit.edu/}{http://places2.csail.mit.edu/}} based on ResNet architecture.

\subsubsection{Image description}
Finally, we use  a caption generator to extract more visual context information from each image, as a natural language description. Image caption generation approaches can use either top-down or bottom-up approaches. The bottom-up approach consists of detecting objects in the image and then attempting to combine the identified objects into a caption  \cite{karpathy2015deep}. On the other hand, the top-down approach learns the semantic representation of the image which is then decoded into the caption \cite{vinyals2015show}. Most current state-of-the-art systems adopt the \textit{top-down} approach using RNN-based architectures. In this work, we use the latter top-down  model to extract the visual description of the image.

The caption generator encoder of  \cite{vinyals2015show} uses a ResNet architecture \cite{he2016deep} trained on ILSVRC competition dataset for general image classification task, and the decoder is tuned on COCO-caption \cite{lin2014microsoft}, the same dataset for which we extract all visual context information.  Table \ref{se:sat-data} shows that the caption has richer semantic.



\begin{figure*}
\centering 

\includegraphics[width=\textwidth]{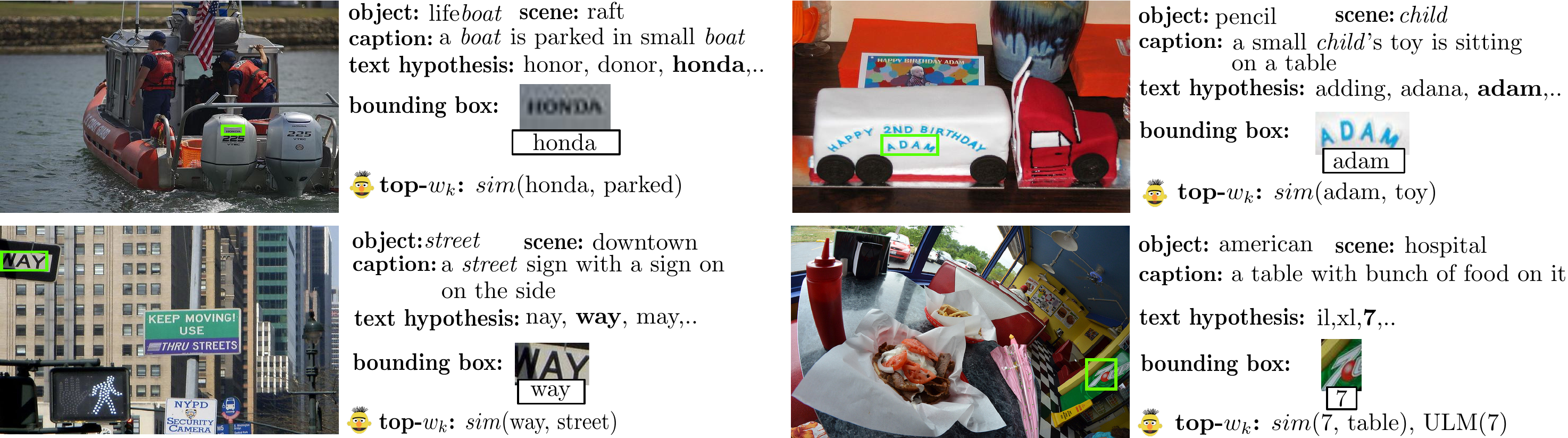}
\vspace*{-0.4cm}
\caption{Examples of our proposed dataset. For each bounding box there are list of text hypotheses $(w_{k})$ and visual context information, object, place and caption. The \textbf{bold} and \textit{italic} font shows the ground-truth and the overlapping visual information, respectively. The top-$w_{k}$ indicates the top re-ranking score based on Bert \cite{devlin2018bert} similarity or Unigram Language Model (ULM). }

 \label{fig:example}
 \end{figure*}

\section{Dataset Construction}

\subsection{Text hypothesis selection}

As described above, the output of several text spotting systems is included in the dataset as text hypotheses or possible candidates for each image.
However, some filtering is applied to remove duplicates and unlikely words:


First, we use a unigram language model (ULM) to filter out rare words (\eg \textit{pretzel}), non-words (\eg \textit{tolin}), or very short words (\eg \textit{inc}) unlikely to be in the image. The ULM \cite{sabir2018visual} was built from Opensubtitles \cite{lison2016opensubtitles2016}\footnote{\href{https://www.opensubtitles.org}{https://www.opensubtitles.org}}, a large database of movie subtitles containing around 3 million unique word forms, including numbers and other alphanumeric combinations that make it well suited for our task. We combined this corpus with google-ngrams\footnote{\href{https://books.google.com/ngrams}{https://books.google.com/ngrams}} that contains 5 million tokens from English literature books. The combined corpora contain around 8 million tokens as shown in Table~\ref{se:Dic}.


Secondly, we add the \textit{ground-truth} if it was removed by the filter or if it was not included in the hypothesis list generated by the baselines. Note that this may occur often, since according to the author of COCO-text \cite{veit2016coco} the significant shortcoming of this dataset is a bounding box detection recall. Therefore, in about 40\% of the images, the text is not properly detected and thus the classifiers fail to recognize it.

\subsection{Visual context selection}
Despite  we extract the top-5 objects from each image, we use a semantic similarity measure and threshold to filter out predictions where the object classifier is not confident enough. We use two approaches to filter out duplicated cases and false positive example.

\vspace{1mm}
\noindent{\bf Threshold measure.} First, we consider a threshold $P(w|class)<0.5$ to extract the most likely classes in the images, and eliminate low confidence predictions.


\vspace{1mm}
\noindent{\bf Semantic alignment.}
We use the cosine similarity to select the most likely visual context in the image. Concretely, we use a general text word-embedding \cite{mikolov2013distributed,pennington2014glove} to compute the similarity score between different visual context elements, and then we select objects or places detected with: 1) a high confidence and that have 2) strong semantic similarity with other image elements. The underlying assumption is that if two objects in the image are related, the classifier prediction we are relying on will be more likely to be correct.  



\begin{table*}[t!]
\centering

\caption{Data statistic for training dataset that publicly available for caption and text spotting. } 
\begin{tabular}{|l|c|c|c|c|c|c|c|c|}

\hline 
\multicolumn{9}{|c|}{Unique Count for Textual Dataset }\\ 
\hline 

 \rowcolor{Gray}
 Dataset & image \# & bbox & caption & object  &words & nouns & verb & adjectives                        \\
\hline 
Conceptual \cite{sharma2018conceptual}   & 3M & -  &3M  & - &34219,055 &     10254,864        &   1043,385      &     3263,654       \\ 
MSCOCO \cite{lin2014microsoft}               & 82k & - & 413k  & -      & 3732,339  & 3401,489             &     250,761             & 424977 \\ 
Flickr 30K \cite{young2014image}  & 30k  & - &160k & - &  2604,646  &   509,459 & 139128  & 169158 \\ 

SVT \cite{wang2010word}        & 350 & $\checkmark$ & - & - &10,437  & 3856   &  46 &   666  \\    
COCO-Text \cite{veit2016coco} & 66k & $\checkmark$ & -& - & 177,547 &  134,970 &  770 & 11,393 \\ 
COCO-Text-V  (ours)    & 16k   & $\checkmark$ &60k  & 120k  & 697,335  &   246,013    &  35,807     & 40,922        \\  
IC17-V  (ours)           & 10k   & $\checkmark$ &25k & 50k  & 296,500  &   96,371     &  15,820     & 15,023         \\ 
COCO-Pairs (ours)         & 66k  &  - & -& 158k &319,178  &  188,295     &   6,878     &     46,983               \\  
 \hline

\end{tabular}
\label{se:sat-data}
\end{table*}

\begin{table}[ht]
\centering

\caption{Total count of unique words - Dictionary.} 
\resizebox{\columnwidth}{!}{
\begin{tabular}{|l|c|c|c|c|}

\hline 
 
\multicolumn{5}{|c|}{Unique Count of Textual Data}\\ 
\hline 
\rowcolor{Gray}
 Dictionary & words  & nouns & verb & adjectives                        \\
 \hline 
Dic-90K \cite{jaderberg2014synthetic}               &   87,629     &   20,146     &  6,956    & 15,534        \\   
ULM \cite{sabir2019semantic}  & 8870,209 & 2695,906 & 139,385 &  824,581  \\  

 \hline

\end{tabular}
}
\label{se:Dic}
\end{table}

\subsection{Object and text co-occurrence database} 

\begin{figure*}[t!]
\centering 
\includegraphics[width=\textwidth]{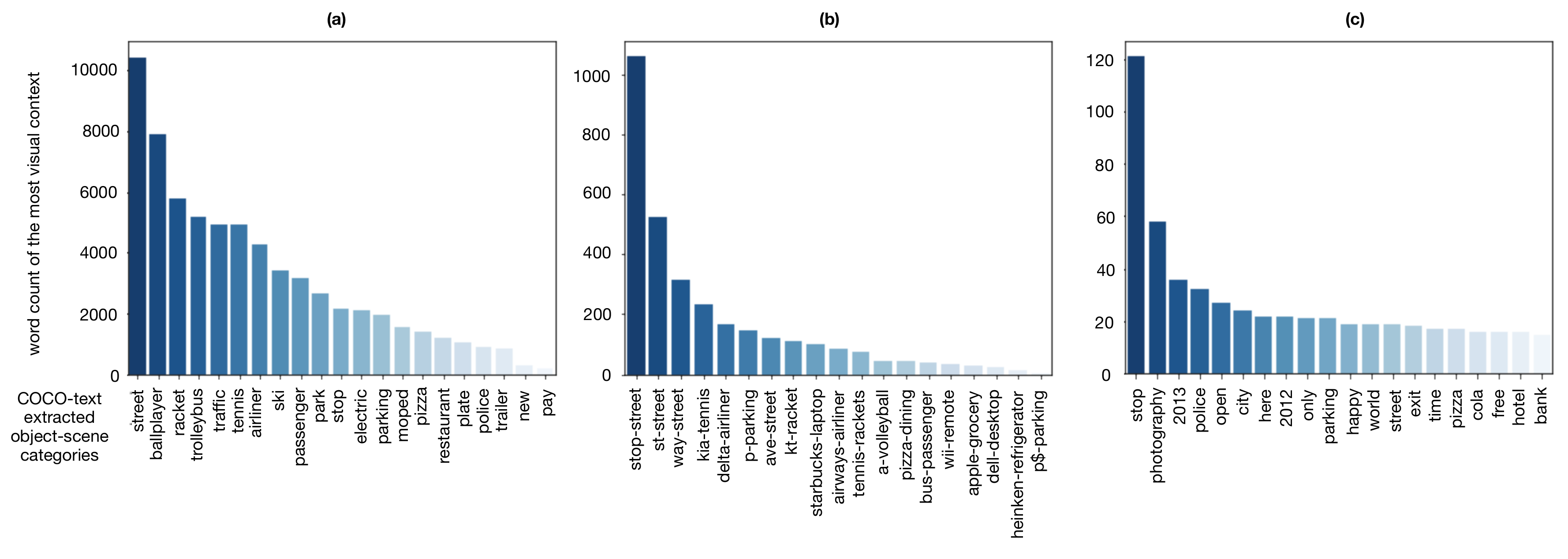}  
\vspace*{-0.6cm}
\caption{(a) Frequency of objects in COCO-text images. (b) Most common pair (text-object) in the training dataset (c) Frequency count of the most visual context in the testing dataset.}
\vspace{-2mm}
\label{fig:all bar}
\end{figure*}

Finally, we enrich the dataset with text-object co-occurrence frequencies. Since this information is not associated to each image, but is an aggregated of the whole dataset, it is provided in a separate table. This co-occurrence information may be useful when the text hypotheses and the scenes are not close in the semantic space but they are in the real world (\eg \textit{delta} and \textit{airliner} or the sports TV channel \textit{kt} and \textit{racket} may not be close according to a general word embedding model, but they co-occur often in the image dataset). A sample of these co-ocurrence frequencies is shown in Figure~\ref{fig:all bar} (b).



 The co-occurrence information \cite{sabir2018visual} can be used to estimate the conditional probability  $P(w \vert c)$ of a word $w$ given that object $c$ appears in the image:
\begin{equation}
\small P(w\vert c)\;=\;\frac{freq(w,c)}{freq(c)}
\end{equation}
where $freq(w,c)$ is the number of training images,  $w$ appears as the gold standard (ground truth) annotation for recognized text, and the object classifier detects object label $c$ in the image. Similarly, $freq(c)$ is the number of training images where the object classifier detects object class $c$.

\subsection{Resulting datasets}
In this section, we outline in more detail our textual visual context dataset, which is an extension to COCO-text. First, we explain the original dataset and then we describe our proposed textual visual context.

\subsubsection{COCO-text without visual context}
As we described in Section \ref{sec:coco-text-explain}, the COCO-text dataset is much larger than other text detection and recognition. It consists of 63,686 images, 173,589 text instances (annotations of the images). 




\subsubsection{COCO-text with visual context}
We propose three different visual textual datasets for COCO-text as shown in Table~\ref{tb:tex-recogition}: 1) training dataset (COCO-Text-V), 2) benchmark testing (IC17-V) and 3) object and text co-occurrence database (COCO-Pairs).

\vspace{1mm}
\noindent{\bf COCO-Text-V:}  It consists of 16K images with associated bounding boxes, and 60K textual data, each line have a caption, object and scene visual information. As shown in Table \ref{se:data}, for each bounding box we extract $k$=10 text hypotheses, and each of them have different or same visual context information depending on the semantic alignment. 

\vspace{1mm}
\noindent{\bf ICDAR17-Task3-V (IC17-V)} is based on ICDAR17 task 3 end-to-end text recognition dataset. Similar to COCO-Text-V, we only introduce the visual context (textual dataset) for each bounding box. It consists of 10K images with 25k textual data for testing and validation.

To be able to use other type of word embedding, knowledge based embedding, we use external knowledge BabelNet \cite{navigli2012babelnet} to extract multiple senses for each word. BabelNet\footnote{\href{https://babelnet.org/}{https://babelnet.org/}} is the largest semantic network with a multilingual encyclopedic dictionary, comprising approximately 16 million entries for named entities linked by semantic relations and concepts. Each class label in  ResNet has sense or meaning that is extracted from the predefined sense inventory (BabelNet). This allows the model to learn more accurate semantic relations between the spotted text and its visual. That sense ID can be used to extract any word vector from any pre-trained sense embedding \cite{iacobacci2015sensembed,pilehvar2016conflated, raganato2016automatic, camacho2016nasari,iacobacci2019lstmembed}. It consists of 1800 images with id senses (\eg orange$_{bn:00059249n}^{1}$ as  \textit{fruit} and orange$_{bn:15347402n}^{2}$ as \textit{color}) that can be used to compute the similarity vector. Some of the words can be used multiple times because they have only one meaning. For example, an ``umbrella''  means the same in all contexts; meanwhile, the word ``bar'' has multiple meanings, such as a steel bar or bar that serves alcoholic beverages.  

\vspace{1mm}
\noindent{\bf COCO-Pairs:}  This textual dataset has no bounding box, only the textual information. The dataset consists of only a pair of object-text extracted from each image.  It consists of 158K word-visual context pairs. We combined the output from the visual classifier with the ground truth to create the pairs (\eg text-scene, text-object).


\begin{figure*}[t]
\centering 

\includegraphics[width=\textwidth]{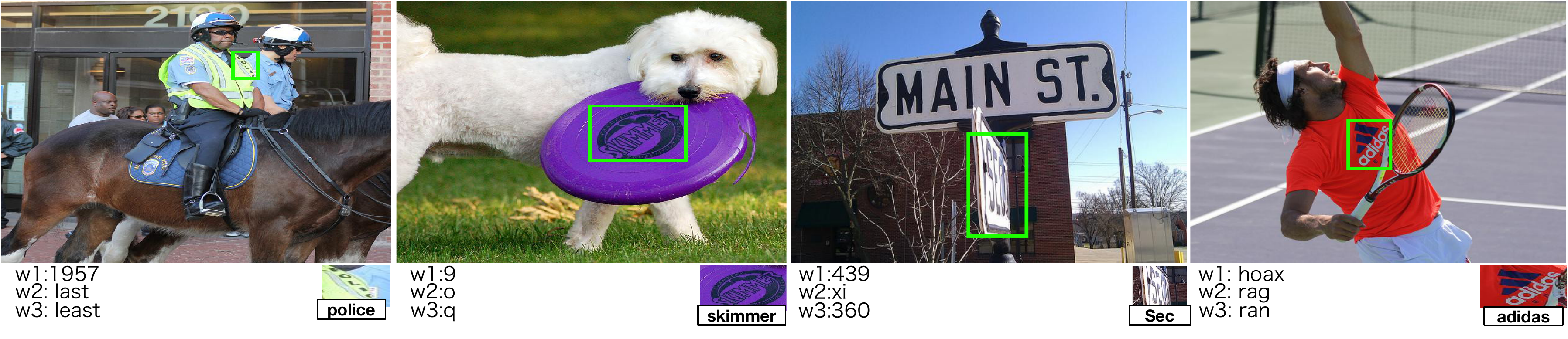} 
\caption{Some random examples extracted from COCO-text with poor detection. The poor detection effect the accuracy of our baseline. Thus, we use the \textbf{ground truth annotation} to overcome this shortcoming in this dataset COCO-text. }
\label{fig:bounding-box}
 \end{figure*} 

Table \ref{se:sat-data} shows unique word count of  part-of-speech tagging (nouns, verb, etc.) of our dataset. Our proposed textual datasets have more semantic than the original   COCO-text dataset. Also, as seen in Figure \ref{fig:bounding-box} real text in the wild is very challenging problem  and thus, current state-of-the-art including our dataset struggle to detect the correct coordination of bounding box. Thus, we use the dataset, COCO-text, ground truth annotation to overcome this shortcoming in this inaccurate bounding box coordination.

 \begin{figure*}[ht]
\centering 
\includegraphics[width=\textwidth]{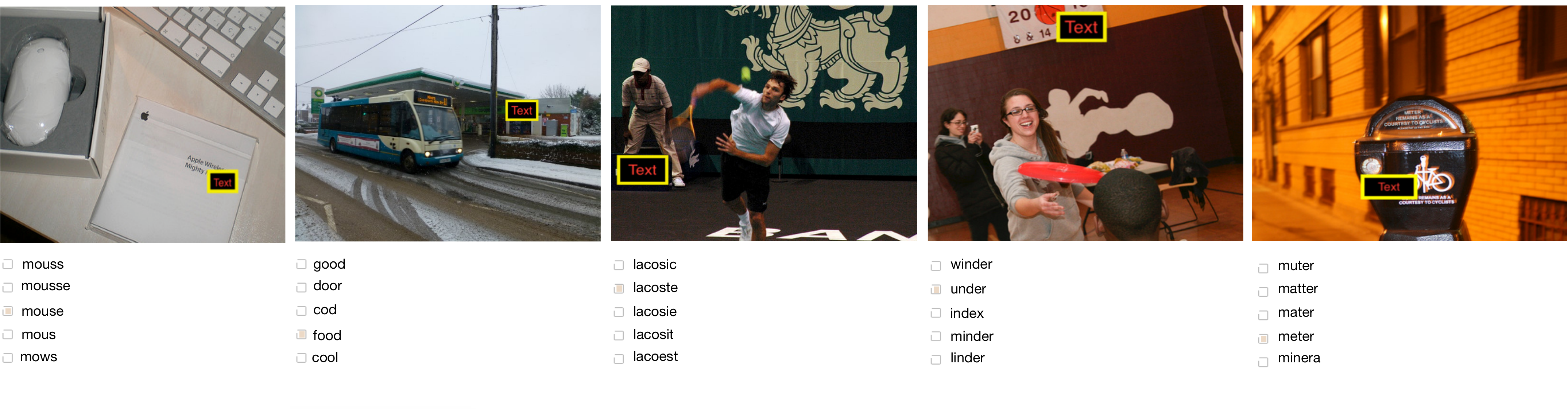} 
\vspace{-7mm}
\caption{The user interface  presented to our human subjects
through the survey website asking them to re-rank the text hypothesis based on the visual information. This figure show samples of the variety of images in the wild in COCO-text such as outdoor and indoor images. In this figure, the $k$=5 text hypothesis has been generated by our baselines and lets the human subject have to choose the most related text to its environmental context.} 
\vspace{-2mm}
 \label{fig:survey}
 \end{figure*} 
 

\section{Experimental Evaluation}

\subsection{Task}
To evaluate the utility of the proposed dataset, we define a novel task, consisting of using the visual context in the image where the text appears to re-rank a list of candidates for the spotted text generated by some pre-existing model. 



More specifically, the task is to use different \textit{similarity} or \textit{relatedness} scorers to reorder the $k$-best hypothesis produced by a trained model with a softmax output. This candidate word  re-ranking should filter out false positive and eliminate low frequency short words. The softmax score and the probabilities of the most related elements in the visual context are then combined by simple algebraic multiplication. In this work, we experimented extracting and re-ranking $k$-best hypotheses for $k=1\ldots10$. 


\subsection{Evaluation remarks} 
For evaluation, we used a less restrictive protocol than the standard one proposed by \cite{wang2010word} and adopted in most state-of-the-art benchmarks, which does not consider words with less than three characters. This protocol was introduced to overcome the false positives on short words that most current state-of-the-art struggle with, including our Baseline. Instead, we consider all cases in the dataset, and words with less than three characters are also evaluated.


Since our task is re-ranking, we use the Mean Reciprocal Rank (MRR) to evaluate the quality of re-ranker outputs. MRR is computed as  $\mathrm{MRR}=\frac{1}{|Q|} \sum_{k=1}^{|Q|} \frac{1}{\operatorname{rank}_{k}}$, where rank $k$ is the position of the first correct answer in the candidate list. MRR is only looking at the rank of the first correct answer; hence it is more suitable in cases such ours, where for each candidate word there is only a single right answer. 

\vspace{1mm}
\noindent{\bf Human Evaluation as an Upper Bound.}
To calibrate the difficulty of the task we picked 33 random pictures from the test dataset and had 16 human subjects try to select the right word among the top $k=5$ candidates produced by the baseline text spotting system. We observed that human subjects more familiar with ads and commercial logos obtain higher scores. Average human performance was 63\% (highest 87\%, lowest 39\%). Figure \ref{fig:survey} shows the user interface for human annotation.



\subsection{Baselines} 
To generate the list of candidate words that will be re-ranked, we rely on two baseline pre-trained systems: a CNN \cite{jaderberg2016reading} and an LSTM \cite{ghosh2017visual}. Each baseline takes a text image bounding box ${\text{Bb}}$ as input and produces $k$ candidate words $w_{1}\ldots{w_k}$ plus a probability for each prediction $P(w_{i}|{\text{Bb}})\:\:\:i=1\ldots{k}$. 

The CNN baseline uses a closed lexicon and can not recognize any word outside its 90K-word dictionary. The LSTM baseline uses a visually soft-attention mechanism which performs unconstrained text recognition without relying on a lexicon.



\subsection{Experiments} 
We performed two experiments, and in each of them we compared the performance of several existing semantic similarity/relatedness systems.

The first experiment consists of re-ranking the text hypotheses produced by the baseline spotting system using only word-to-word similarity metrics. In this experiment each candidate word is compared to objects and places appearing in the image, and re-ranked according to the obtained similarity scores.
In the second experiment, we re-rank the candidate words comparing them with an automatically generated caption for the image. For this, we require semantic similarity systems able to produce word-to-sentence or sentence-to-sentence similarity scores.

\subsubsection{Experiment 1: Re-ranking using word-to-sentence metrics}
We used different off-the-shelf semantic similarity systems to compare the candidate words with the visual context in the image (objects and places), and evaluated the performance of each of them.  The used systems are:

\vspace{1mm}
\noindent{\bf Glove \cite{pennington2014glove}:}  Word
embedding system that derives the semantic relationships between words from the co-occurrence matrix. The advantage of Glove over Word2Vec \cite{mikolov2013distributed} is that it does not rely on local word-context information, but it incorporates global co-occurrence statistics.



\vspace{1mm}
\noindent{\bf Fasttext \cite{joulin2017bag}:}  Extension of Word2Vec  that instead of learning   directly the   word, it learns a $n$-gram  representation. Thus, it can deal with rare words not seen during training, by breaking them down into character $n$-grams.  

\vspace{1mm}
\noindent{\bf Relational Word Embeddings \cite{camacho2019relational} (RWE): } Enhanced version of Word2Vec that encodes complementary relational knowledge into the standard word-embedding in the semantic space. This enhanced embedding is still learned from pure co-occurrence statistics and not relying on any external knowledge. The model intends to capture and combine new knowledge complementary to standard similarity-centric embeddings.

\vspace{1mm}
\noindent{\bf TWE \cite{sabir2018visual}:}  Semantic Relatedness with Word Embeddings. Word embedding trained using Word2Vec, but instead of general corpus, it is trained on the presented dataset, so it can learn associations between candidate words and their visual context that are uncommon in general text. The model is trained on a  \textit{Skip-gram} model \cite{mikolov2013distributed} that works well with small amounts of training data and is able to represent low-frequency words. 

\vspace{1mm}
\noindent{\bf LSTMEmbed \cite{iacobacci2019lstmembed}:}  LSTMEmbed is the most recent model in sense embeddings. It utilizes a BiLSTM   architecture to learn the word and sense embeddings from annotated corpora.  We use the same approach than in  \cite{iacobacci2019lstmembed}: 200-dimension embeddings trained on the English portion of BabelWiki and English Wikipedia.

Once the similarity between the candidate word and the most closely related element in the  visual context is computed, we need to convert that score to a probability in order to combine them in the re-ranking process. Following \cite{sabir2018visual}, we use two different methods to  obtain the final probability:

\begin{itemize}
\setlength\itemsep{-0.5em}   
    \item  For TWE, we use $\operatorname{P}_{TWE}(w | c)=\frac{\tanh (\operatorname{sim}(w, c))+1}{2 P(c)}$ where, since $\tanh(x) \in [-1,1]$, then $\tanh(x)+1 \in [0,2]$, and thus $\frac{\tanh(x)+1}{2} \in [0,1]$ is our approximation of $P(w,c)$, which is then divided by $P(c)$ to obtain the conditional probability.
    \vspace{1mm}
    \item  For all other word-level similarity methods, we combine the obtained cosine similarity $sim(w,c)$,  the probability $P(c)$ of the detected context (provided by the object/place classifier), and the probability $P(w)$ of the candidate word (estimated from a 8M token corpus \cite{lison2016opensubtitles2016}). The final probability is computed following  \cite{blok2003probability}  with confirmation assumption $p(w | c) \geqslant p(w)$ as:
\begin{equation*}
\label{eq:sim}
P(w\vert c)=P(w)^\alpha \text{~~~~where~} \alpha=\left({\textstyle\frac{1-sim(w,c)}{1+sim(w,c)}}\right)^{1-P(c)\;}
\end{equation*}
\end{itemize}
 
Results of experiment 1 are shown in Table~\ref{table_results}-top.  



\subsubsection{Experiment 2: Re-ranking using word-to-sentence metrics}
In the second experiment we used sentence-level semantic similarity. For this, we resorted to state-of-the-art sentence embedding models fine-tuned using the caption dataset.


\vspace{1mm}
\noindent{\bf USE-Transformer \cite{cer2018universal}:}  Universal Sentence Encoder (USE) is the current state-of-the-art in Semantic Textual Similarity (STS). The model  is based on the transformer architecture USE-T \cite{vaswani2017attention} that targets high accuracy at the cost of complexity and resource consumption. We experimented with USE-T fine tuning and feature extraction to compute the semantic relation with cosine distance.



\vspace{1mm}
\noindent{\bf Bert\footnote{We use the basic bert-base-uncased model.} \cite{devlin2018bert}:}  Bidirectional Encoder Representations from Transformers has shown groundbreaking results in many semantics-related NLP tasks.

\vspace{1mm}
\noindent{\bf Fine-tuned Bert:}
According to Bert authors, it is not suited for Semantic Textual Similarity (STS) task, since it does not generate a meaningful vector to compute the cosine distance. Thus, we also evaluated a fine-tuned version of the model with one extra layer to compute the semantic score between caption and candidate word. In particular, we fed the sentence representation into a linear layer and a softmax for sentence pair tasks (Q\&A re-ranking task). 
\medskip


Results for the second experiment are shown in Table~\ref{table_results}-bottom. Fine-tuned Bert outperforms all other models. BL+TWE ranks second in accuracy. 

\begin{table}[t!]
\caption{Experimental results. Row BL shows the baseline performance, without any visual context information. Gray-shaded indicate the models has been trained  or fine-tuned using the presented dataset. Star  $\bigstar$ indicate that the model relies on a  predefined sense inventory and annotated data. Accuracy (\textit{Acc.}) is the percentage of images in which the right word is ranked in the first place. Column $k$ shows the number of $k$-best hypotheses re-ranked to obtain the shown accuracy. MRR is computed using $k=8$ for CNN and $k=4$ for LSTM.}




\centering
\resizebox{\columnwidth}{!}{
\begin{tabular}{|l|c c | c  | c c | c |} 

\hline 
 \textbf{Model}  & \multicolumn{3}{c|}{\textbf{CNN}}   & \multicolumn{3}{c|}{\textbf{LSTM}}  \\ 
  
  & \textit{Acc.} &   \textit{k} &  MRR &   \textit{Acc.} &   \textit{k} & MRR \\ 

\hline
\hline
Baseline (BL)  &  \multicolumn{3}{c|}{\textbf{Acc.:19.7}}   & \multicolumn{3}{c|}{\textbf{Acc.:17.9}}  \\




\hline \hline
\multicolumn{7}{|l|}{\bf Experiment 1}\\
\hline 
BL+Word2vec   \cite{mikolov2013distributed}    &   21.8        &  5   & 44.3 & 19.5        & 4     &80.4     \\  

BL+ Glove       \cite{pennington2014glove}    &   22.0         &  7   & 44.5    & 19.1            & 4  & 78.8               \\


BL+Sw2v \cite {mancini2017embedding}  $\bigstar$  &  21.8        & 7    &  44.3  & 19.4         & 4     &80.1     \\ 

BL+Fasttext \cite{joulin2017bag}  &   21.9            &  7   & 44.6 & 19.4               & 4       & 80.3       \\ 

\rowcolor{Gray}
BL+TWE  \cite{sabir2018visual}   &   22.2         &  7   & 44.7 & 19.5     & 4     &80.2   \\


BL+RWE \cite{camacho2019relational} &   21.9      &  7   &44.5 & 19.6        & 4     &80.7     \\

BL+ LSTMmebed \cite{iacobacci2019lstmembed} $\bigstar$  &   21.6   &     7   & 44.0   & 19.2        & 4     & 79.6     \\

\hline 
\hline 
\multicolumn{7}{|l|}{\bf Experiment 2}\\
\hline 


\rowcolor{Gray}
BL+USE-T \cite{cer2018universal}                &   22.0         &  6  & 44.7   & 19.2      & 4    & 79.5       \\ 


BL+ BERT-feature   \cite{devlin2018bert}     &   21.7       &    7 & 45.0    & 19.3             &   4 &  \textbf{81.2}                     \\

\rowcolor{Gray}
BL+ BERT (fine-tune)   \cite{devlin2018bert}     &   \textbf{22.7}         &    8 & \textbf{45.9}    & \textbf{20.1}             &   9 &  79.1        \\

\hline 

\end{tabular}
}
\label{table_results}
\end{table}




\section{Conclusions and Further Work}

We have proposed a dataset that extends COCO-text with visual context information, that we believe useful for the text spotting problem.  In contrast to the most recent method \cite{Prasad_2018_ECCV} that relies on limited classes of context objects and uses a complex architecture to extract visual information, our approach utilizes out-of-the-box state-of-the-art tools. Therefore, the  dataset annotation will  be improved in the future as better systems become available. This dataset can be used to leverage semantic relation between image context and candidate texts into text spotting systems, either as post-processing or end-to-end training. We also use our dataset to train/tune an evaluate existing semantic similarity systems when applied to the task of re-ranking text hypothesis produced by a text spotting baseline, showing that it can improve the accuracy of the original 
baseline between 2 and 3 points. Note that there's a lot of room for improvement up to 7.4 points in a benchmark dataset.

 \section*{Acknowledgments}
This work is supported by the KASP Scholarship Program and by the Spanish government under projects HuMoUR TIN2017-90086-R and   Mar\'ia de Maeztu Seal of Excellence MDM-2016-0656.




{\small
\bibliographystyle{ieee_fullname}
\bibliography{egbib}

\begin{thebibliography}{10}\itemsep=-1pt

\bibitem{bissacco2013photoocr}
Alessandro Bissacco, Mark Cummins, Yuval Netzer, and Hartmut Neven.
\newblock Photoocr: Reading text in uncontrolled conditions.
\newblock In {\em CVPR}, 2013.

\bibitem{blok2003probability}
Sergey Blok, Douglas Medin, and Daniel Osherson.
\newblock Probability from similarity.
\newblock In {\em AAAI}, 2003.

\bibitem{camacho2019relational}
Jose Camacho, Luis Espinosa-Anke, and Steven Schockaert.
\newblock Relational word embeddings.
\newblock {\em arXiv preprint arXiv:1906.01373}, 2019.

\bibitem{camacho2016nasari}
Jos{\'e} Camacho-Collados, Mohammad~Taher Pilehvar, and Roberto Navigli.
\newblock Nasari: Integrating explicit knowledge and corpus statistics for a
  multilingual representation of concepts and entities.
\newblock {\em Artificial Intelligence}, 2016.

\bibitem{cer2018universal}
Daniel Cer, Yinfei Yang, Sheng-yi Kong, Nan Hua, Nicole Limtiaco, Rhomni~St
  John, Noah Constant, Mario Guajardo-Cespedes, Steve Yuan, Chris Tar, et~al.
\newblock Universal sentence encoder.
\newblock {\em arXiv preprint arXiv:1803.11175}, 2018.

\bibitem{devlin2018bert}
Jacob Devlin, Ming-Wei Chang, Kenton Lee, and Kristina Toutanova.
\newblock Bert: Pre-training of deep bidirectional transformers for language
  understanding.
\newblock {\em arXiv preprint arXiv:1810.04805}, 2018.

\bibitem{fang2018attention}
Shancheng Fang, Hongtao Xie, Zheng-Jun Zha, Nannan Sun, Jianlong Tan, and
  Yongdong Zhang.
\newblock Attention and language ensemble for scene text recognition with
  convolutional sequence modeling.
\newblock In {\em ACMMM}, 2018.

\bibitem{ghosh2017visual}
Suman~K Ghosh, Ernest Valveny, and Andrew~D Bagdanov.
\newblock Visual attention models for scene text recognition.
\newblock {\em arXiv preprint arXiv:1706.01487}, 2017.

\bibitem{gomez2017icdar2017}
Raul Gomez, Baoguang Shi, Lluis Gomez, Lukas Numann, Andreas Veit, Jiri Matas,
  Serge Belongie, and Dimosthenis Karatzas.
\newblock Icdar2017 robust reading challenge on coco-text.
\newblock In {\em ICDAR}, 2017.

\bibitem{graves2006connectionist}
Alex Graves, Santiago Fern{\'a}ndez, Faustino Gomez, and J{\"u}rgen
  Schmidhuber.
\newblock Connectionist temporal classification: labelling unsegmented sequence
  data with recurrent neural networks.
\newblock In {\em ICML}, 2006.

\bibitem{he2016deep}
Kaiming He, Xiangyu Zhang, Shaoqing Ren, and Jian Sun.
\newblock Deep residual learning for image recognition.
\newblock In {\em CVPR}, 2016.

\bibitem{hochreiter1997long}
Sepp Hochreiter and J{\"u}rgen Schmidhuber.
\newblock Long short-term memory.
\newblock {\em Neural computation}, 1997.

\bibitem{iacobacci2019lstmembed}
Ignacio Iacobacci.
\newblock Lstmembed: Learning word and sense representations from a large
  semantically annotated corpus with long short-term memories.
\newblock In {\em ACL}, 2019.

\bibitem{iacobacci2015sensembed}
Ignacio Iacobacci, Mohammad~Taher Pilehvar, and Roberto Navigli.
\newblock Sensembed: Learning sense embeddings for word and relational
  similarity.
\newblock In {\em ACL}, 2015.

\bibitem{jaderberg2014synthetic}
Max Jaderberg, Karen Simonyan, Andrea Vedaldi, and Andrew Zisserman.
\newblock Synthetic data and artificial neural networks for natural scene text
  recognition.
\newblock {\em arXiv preprint arXiv:1406.2227}, 2014.

\bibitem{jaderberg2016reading}
Max Jaderberg, Karen Simonyan, Andrea Vedaldi, and Andrew Zisserman.
\newblock Reading text in the wild with convolutional neural networks.
\newblock {\em IJCV}, 2016.

\bibitem{joulin2017bag}
Armand Joulin, Edouard Grave, and Piotr Bojanowski~Tomas Mikolov.
\newblock Bag of tricks for efficient text classification.
\newblock {\em EACL}, 2017.

\bibitem{kang2017detection}
Chulmoo Kang, Gunhee Kim, and Suk~I Yoo.
\newblock Detection and recognition of text embedded in online images via
  neural context models.
\newblock In {\em AAAI}, 2017.

\bibitem{karatzas2013icdar}
Dimosthenis Karatzas, Faisal Shafait, Seiichi Uchida, Masakazu Iwamura,
  Lluis~Gomez i Bigorda, Sergi~Robles Mestre, Joan Mas, David~Fernandez Mota,
  Jon~Almazan Almazan, and Lluis~Pere de~las Heras.
\newblock Icdar 2013 robust reading competition.
\newblock In {\em ICDAR}, 2013.

\bibitem{karpathy2015deep}
Andrej Karpathy and Li Fei-Fei.
\newblock Deep visual-semantic alignments for generating image descriptions.
\newblock In {\em CVPR}, 2015.

\bibitem{liambas2016autonomous}
Christos Liambas and Miltiadis Saratzidis.
\newblock Autonomous ocr dictating system for blind people.
\newblock In {\em GHTC}, 2016.

\bibitem{lin2014microsoft}
Tsung-Yi Lin, Michael Maire, Serge Belongie, James Hays, Pietro Perona, Deva
  Ramanan, Piotr Doll{\'a}r, and C~Lawrence Zitnick.
\newblock Microsoft coco: Common objects in context.
\newblock In {\em ECCV}, 2014.

\bibitem{lison2016opensubtitles2016}
Pierre Lison and J{\"o}rg Tiedemann.
\newblock {O}pen{S}ubtitles2016: Extracting large parallel corpora from movie
  and {TV} subtitles.
\newblock In {\em LREC}, 2016.

\bibitem{mancini2017embedding}
Massimiliano Mancini, Jose Camacho-Collados, Ignacio Iacobacci, and Roberto
  Navigli.
\newblock Embedding words and senses together via joint knowledge-enhanced
  training.
\newblock In {\em CoNLL}, 2017.

\bibitem{mikolov2013distributed}
Tomas Mikolov, Ilya Sutskever, Kai Chen, Greg~S Corrado, and Jeff Dean.
\newblock Distributed representations of words and phrases and their
  compositionality.
\newblock In {\em NeurIPS}, 2013.

\bibitem{navigli2012babelnet}
Roberto Navigli and Simone~Paolo Ponzetto.
\newblock Babelnet: The automatic construction, evaluation and application of a
  wide-coverage multilingual semantic network.
\newblock {\em Artificial Intelligence}, 2012.

\bibitem{pennington2014glove}
Jeffrey Pennington, Richard Socher, and Christopher Manning.
\newblock Glove: Global vectors for word representation.
\newblock In {\em EMNLP}, 2014.

\bibitem{pilehvar2016conflated}
Mohammad~Taher Pilehvar and Nigel Collier.
\newblock De-conflated semantic representations.
\newblock {\em arXiv preprint arXiv:1608.01961}, 2016.

\bibitem{Prasad_2018_ECCV}
Shitala Prasad and Adams Wai Kin~Kong.
\newblock Using object information for spotting text.
\newblock In {\em ECCV}, 2018.

\bibitem{priambada2017levensthein}
Satria Priambada and Dwi~H Widyantoro.
\newblock Levensthein distance as a post-process to improve the performance of
  ocr in written road signs.
\newblock In {\em ICIC}, 2017.

\bibitem{raganato2016automatic}
Alessandro Raganato, Claudio~Delli Bovi, and Roberto Navigli.
\newblock Automatic construction and evaluation of a large semantically
  enriched wikipedia.
\newblock In {\em IJCAI}, 2016.

\bibitem{Ramisa_pami2017}
Arnau Ramisa, Fei Yan, Francesc Moreno-Noguer, and Krystian Mikolajczyk.
\newblock Breakingnews: Article annotation by image and text processing.
\newblock {\em TPAMI}, 2017.

\bibitem{sabir2019semantic}
Ahmed Sabir, Francesc Moreno, and Llu{\'\i}s Padr{\'o}.
\newblock Semantic relatedness based re-ranker for text spotting.
\newblock In {\em EMNLP}, 2019.

\bibitem{sabir2018visual}
Ahmed Sabir, Francesc Moreno-Noguer, and Llu{\'\i}s Padr{\'o}.
\newblock Visual re-ranking with natural language understanding for text
  spotting.
\newblock In {\em ACCV}, 2018.

\bibitem{sharma2018conceptual}
Piyush Sharma, Nan Ding, Sebastian Goodman, and Radu Soricut.
\newblock Conceptual captions: A cleaned, hypernymed, image alt-text dataset
  for automatic image captioning.
\newblock In {\em ACL}, 2018.

\bibitem{shi2016end}
Baoguang Shi, Xiang Bai, and Cong Yao.
\newblock An end-to-end trainable neural network for image-based sequence
  recognition and its application to scene text recognition.
\newblock {\em TPAMI}, 2016.

\bibitem{szegedy2017inception}
Christian Szegedy, Sergey Ioffe, Vincent Vanhoucke, and Alexander~A Alemi.
\newblock Inception-v4, inception-resnet and the impact of residual connections
  on learning.
\newblock In {\em AAAI}, 2017.

\bibitem{szegedy2015going}
Christian Szegedy, Wei Liu, Yangqing Jia, Pierre Sermanet, Scott Reed, Dragomir
  Anguelov, Dumitru Erhan, Vincent Vanhoucke, and Andrew Rabinovich.
\newblock Going deeper with convolutions.
\newblock In {\em CVPR}, 2015.

\bibitem{vaswani2017attention}
Ashish Vaswani, Noam Shazeer, Niki Parmar, Jakob Uszkoreit, Llion Jones,
  Aidan~N Gomez, {\L}ukasz Kaiser, and Illia Polosukhin.
\newblock Attention is all you need.
\newblock In {\em NeurIPS}, 2017.

\bibitem{veit2016coco}
Andreas Veit, Tomas Matera, Lukas Neumann, Jiri Matas, and Serge Belongie.
\newblock Coco-text: Dataset and benchmark for text detection and recognition
  in natural images.
\newblock {\em arXiv preprint arXiv:1601.07140}, 2016.

\bibitem{vinyals2015show}
Oriol Vinyals, Alexander Toshev, Samy Bengio, and Dumitru Erhan.
\newblock Show and tell: A neural image caption generator.
\newblock In {\em CVPR}, 2015.

\bibitem{wang2010word}
Kai Wang and Serge Belongie.
\newblock Word spotting in the wild.
\newblock {\em Computer Vision--ECCV}, 2010.

\bibitem{xu2015show}
Kelvin Xu, Jimmy Ba, Ryan Kiros, Kyunghyun Cho, Aaron Courville, Ruslan
  Salakhudinov, Rich Zemel, and Yoshua Bengio.
\newblock Show, attend and tell: Neural image caption generation with visual
  attention.
\newblock In {\em ICML}, 2015.

\bibitem{young2014image}
Peter Young, Alice Lai, Micah Hodosh, and Julia Hockenmaier.
\newblock From image descriptions to visual denotations: New similarity metrics
  for semantic inference over event descriptions.
\newblock {\em TACL}, 2014.

\bibitem{zhou2017places}
Bolei Zhou, Agata Lapedriza, Aditya Khosla, Aude Oliva, and Antonio Torralba.
\newblock Places: A 10 million image database for scene recognition.
\newblock {\em TPAMI}, 2017.

\bibitem{zhu2016could}
Anna Zhu, Renwu Gao, and Seiichi Uchida.
\newblock Could scene context be beneficial for scene text detection?
\newblock {\em Pattern Recognition}, 2016.

\end{thebibliography}
}

\end{document}